# Deep Learning-Based Diffusion MRI Tractography: Integrating Spatial and Anatomical Information

Yiqiong Yang[a,b], Yitian Yuan[a,b], Baoxing Ren[a,b], Ye Wu[c], Yanqiu Feng[a,b,d,e], Xinyuan Zhang[a,b*]

[a] School of Biomedical Engineering, Southern Medical University, Guangzhou 510515, China
[b] Guangdong Provincial Key Laboratory of Medical Image Processing and Guangdong Province Engineering Laboratory for Medical Imaging and Diagnostic Technology, Southern Medical University, Guangzhou 510515, China
[c] School of Computer Science and Technology, Nanjing University of Science and Technology, Nanjing, China
[d] Guangdong-Hong Kong-Macao Greater Bay Area Center for Brain Science and Brain-Inspired Intelligence & Key Laboratory of Mental Health of the Ministry of Education, Southern Medical University, Guangzhou, China
[e] Department of Radiology, Shunde Hospital, Southern Medical University (The First People's Hospital of Shunde, Foshan), Foshan, China
[*] Correspondence to Xinyuan Zhang.

**Corresponding Author:**
Xinyuan Zhang
School of Biomedical Engineering, Southern Medical University
Guangzhou 510515, China.
E-mail address: zhangxyn@smu.edu.cn (X. Zhang).

# Deep Learning-Based Diffusion MRI Tractography: Integrating Spatial and Anatomical Information

**Abstract**

Diffusion MRI tractography technique enables non-invasive visualization of the white matter pathways in the brain. It plays a crucial role in neuroscience and clinical fields by facilitating the study of brain connectivity and neurological disorders. However, the accuracy of reconstructed tractograms has been a longstanding challenge. Recently, deep learning methods have been applied to improve tractograms for better white matter coverage, but often comes at the expense of generating excessive false-positive connections. This is largely due to their reliance on local information to predict long-range streamlines. To improve the accuracy of streamline propagation predictions, we introduce a novel deep learning framework that integrates image-domain spatial information and anatomical information along tracts, with the former extracted through convolutional layers and the later modeled via a Transformer-decoder. Additionally, we employ a weighted loss function to address fiber class imbalance encountered during training. We evaluate the proposed method on the simulated ISMRM 2015 Tractography Challenge dataset, achieving a valid streamline rate of 66.2%, white matter coverage of 63.8%, and successfully reconstructing 24 out of 25 bundles. Furthermore, on the multi-site Tractoinferno dataset, the proposed method demonstrates its ability to handle various diffusion MRI acquisition schemes, achieving a 5.7% increase in white matter coverage and a 4.1% decrease in overreach compared to RNN-based methods.

# 1. Introduction

Diffusion Magnetic Resonance Imaging (diffusion MRI)-based tractography enables the noninvasive reconstruction and visualization of white matter pathways in the brain. The mapping of structural connectivity and tract-specific quantitative analysis are both based on tractography technique. Due to its critical role, tractography is widely used in research and clinical applications to understand brain development and disease progression.

Tractography typically begins by initiating tracking from selected seed points, with streamlines iteratively growing along the propagation direction. In conventional methods, a predefined map of local fiber orientations—such as the diffusion tensor (DT)(Basser et al., 1994) map or fiber orientation distribution function (fODF)(Tournier et al., 2007; Tournier et al., 2004) map derived from diffusion MRI data—is used to guide the propagation. Three main types of methods are then available to propagate streamline sprouts iteratively until termination criteria are met, ultimately reconstructing the complete fiber tracts. Among them, deterministic propagation methods(Mori et al., 1999) iteratively move along the direction that aligns with one of the peak fODF directions. Probabilistic methods(Behrens et al., 2003) select propagation directions from a distribution of possible orientations to form streamlines. Unlike deterministic and probabilistic methods, Global methods(Reisert et al., 2011) reconstruct all streamlines simultaneously through the optimization of a global loss. Although well-established and widely used, these methods rely on predefined diffusion models, that struggle in regions with complex fiber configurations, such as narrow intersections, bottlenecks, kissing fibers (Rheault et al., 2020).

Recently, numerous neural network-based methods have been proposed to improve tractography by leveraging powerful feature extraction and non-linear mapping capabilities. Unlike conventional methods, these approaches such as Learn-To-Track(Poulin et al., 2017), Entrack(Wegmayr and Buhmann, 2021), DeepTract(Benou and Riklin Raviv, 2019) and the one by Neher et al. (Neher et al., 2017), directly predict the propagation directions from raw diffusion MRI data, eliminating the need for predefined diffusion model. Among these, Learn-to-Track(Poulin et al., 2017) utilizes recurrent neural network to reconstruct streamlines, and Entrack(Wegmayr et al., 2018) estimates maximum entropy posteriors of local streamline propagation directions with machine learning methods, both of which resulting in enhanced white matter coverage and a reduced false positive rate. Some other approaches, for example GESTA(Legarreta et al., 2023), generates complete streamlines in a single step rather than iterative propagation and achieves high white matter coverage. Track-to-Learn(Théberge et al., 2021) and TractOracle(Théberge et al., 2024), two deep reinforcement learning-based methods, remove the need for mutually curated reference streamlines and demonstrate generalizability to unseen data.

Despite these advancements, most DL-based approaches, like traditional methods, rely heavily on local information (i.e., the immediate position being predicted) to guide streamline propagation. Many studies (Maier-Hein et al., 2017; Rheault et al., 2020; Schilling et al., 2019b) suggest such reliance prevents overcoming the specificity-sensitivity curse, particularly in areas with complex fiber configurations where multiple orientations coexist. In these regions, such methods treat all plausible directions as equally likely for distinct trajectories based on only local information, often resulting in the generation of erroneous streamlines (false positives) and the omission of true fiber pathways (false negatives), even when local fiber population orientations are accurately resolved. This highlights the critical need for integrating contextual information to improve the reconstruction of streamlines across the whole brain.

Building upon this limitation, Learn-To-Track(Poulin et al., 2017) leverages recurrent neural networks (RNNs) to retain information from previous streamline segments, enabling the integration of past trajectory information during tracking. However, RNNs are inherently limited in their ability to capture long-range dependencies. It restricts their effectiveness in modeling the global structure of streamlines, which represent the white matter pathways that span across the brain. Attention mechanisms, in contrast, have proven to be highly effective in tasks requiring long-range contextual understanding, making them a promising candidate for addressing the challenges of long-range information integration in streamline tractography. At the same time, image-domain neighborhood information provides crucial spatial context from the surrounding image region, helping to resolve local ambiguities, such as multiple fibers coexisting within a single voxel. While global contextual information captures the overall trajectory of streamlines, neighborhood information focuses specifically on disambiguating these complex local structures, as demonstrated in previous studies(Neher et al., 2017; Poulin et al., 2017; Wegmayr and

Buhmann, 2021).

Given these insights, we propose a deep learning-based tractography framework that combines a convolutional neural network (CNN) for image-domain feature extraction and a decoder-only Transformer model for integrating global contextual information along streamlines, aimed at improving the accuracy and reliability of tractography. Our main contributions are as follow:

1. We propose a novel DL-based framework that simultaneously integrates image-domain neighborhood information as well as global contextual information along streamlines.

2. We design a decoder-only Transformer model based on Generative Pre-trained Transformer (GPT) to effectively generate streamlines, leveraging its capability to handle long-range dependencies.

3. We perform an ablation study to validate each type of information we used.

4. We demonstrate competitive results when compared to state-of-the-art deep learning and classical tractography algorithms, achieving robust performance across datasets from different centers.

## 2. Methods

We introduce a multi-information-aided tractography method leveraging a convolutional block and Transformer-decoder blocks. The overall pipeline of the proposed framework is illustrated in **Fig. 1**(a), where the network predicts propagation directions iteratively to reconstruct streamlines, as detailed in Section 2.1. Meanwhile, the architecture of the network, including the convolutional block and Transformer-decoder block, is shown in **Fig. 1**(b) and (c) and further explained in Section 2.2.

*2.1 Overview of the Proposed Tractography Pipeline*

To accommodate various acquisition schemes, the raw diffusion MRI signal is represented using the coefficients of spherical harmonics (SH) expansion. In our experiments, the maximum harmonics order ($l_{max}$) is set to 6, corresponding to 28 terms in the spherical harmonic expansion of DWI.

A set of $N$ streamlines is represented as $S = \{S_1, S_2, \ldots, S_N\}$. Each streamline is stored as an ordered sequence of coordinates, $S_i = [s_i^1, s_i^2, \ldots, s_i^t]$, where $t$ is the number of vertices of streamline $S_i$, $t_{max} = T$, $s_i^j \in R^3$ denotes the location of the $j_{th}$ vertex along the streamline $S_i$. The distance between two adjacent vertices is called step size $\alpha$. When tracking procedure starts from a seed point or a partially known streamline of $S_i$(incomplete), our algorithm first extracts the corresponding local and neighbourhood diffusion MRI signals along the incomplete streamline $S_i$, resulting in the input sequence $(x_1, \ldots, x_t)$. Subsequently, the proposed network works to encode neighbourhood information and attends to the contextual information along the sequence to predict the propagation direction $\hat{y}_t$. The streamline then grows iteratively based on the specified step size and the predicted propagation direction, according to $s_i^{t+1} = s_i^t + \alpha \cdot \hat{y}_t$. The updated streamline (incomplete) will be the new input to our method and the process is repeated iteratively until the stopping criterion is met. Our model is trained to minimize the weighted mean squared-error loss between the predicted direction of propagation $\hat{y}$ and the reference direction $y$.

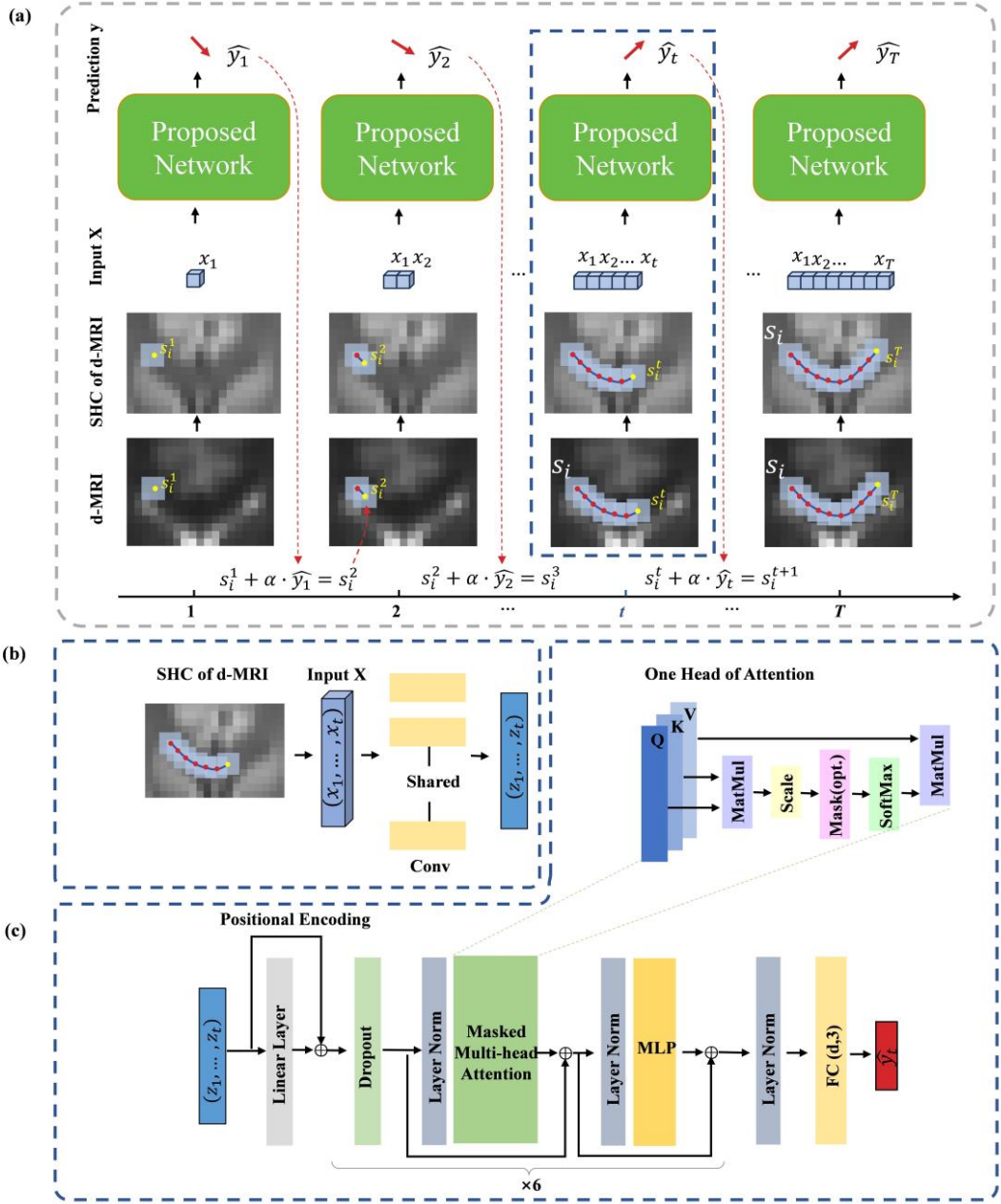

**Fig. 1.** (a) Overview of our proposed tractography algorithm: Given a seed point or a partially known streamline, our method extracts the corresponding local and neighbourhood dMRI signals of it to form the input data sequence $(x_1,...,x_t)$. This sequence is then fed to our network to predict the direction of propagation. Subsequently, the streamline grows based on the given step size and the propagation direction. The updated streamline (incomplete) will be the new input to our method and the process is repeated iteratively until the stopping criterion is met. (b) The process of integrating image-domain information. (c) Architecture of the block used to embed positional information and context information, and detailed illustration of one head self-attention mechanism used in our work.

*2.2 Proposed Network*

*2.2.1 Incorporating Image-domain Neighbourhood Information*

As illustrated in **Fig. 1**(a) and (b), for each vertex of the streamline $S_i$, the signals from the vertex itself and its 3×3×3 neighbourhood is extracted to form the corresponding data sequence $(x_1,...,x_t)$, where $x_t \in R^{3\times3\times3\times28}$. A convolutional layer with a 3×3×3 kernel is applied to encode this image-domain neighbourhood information to a sequence of continuous representations $(z_1,...,z_t)$, where $z_t \in R^{1\times192}$, 192 is the feature dimension. The set of filters learned by the convolutional layer enables the extraction of patterns and preservation of spatial coherence from the neighbourhood, enabling the newly generated sequence encapsulate underlying structures.

*2.2.2 Incorporating Contextual Information*

The anatomical contextual information aggregation block takes the sequences $(z_1,...,z_t)$ produced by the convolution layer as input and generates $\hat{y}_t$ for every vertex. It consists of two main components: the positional encoding layer and the Transformer-decoder stacks(Vaswani et al., 2017). A detailed representation of this block is shown in **Fig. 1**(c).

The positional encoding layer aims to incorporate information about the relative position of the vertices in the sequence, as the streamlines' geometry is closely related to the order of vertices. The learned positional encodings have the same dimension as the input embeddings and are added on them. Then the resultant sequence is fed to the masked self-attention block. This block computes three embedding matrices: the query matrix $Q$, the key matrix $K$, and the value matrix $V$, each size of $t \times 192$. The output is computed as a weighted sum of the values, where the weight assigned to each value is computed by a compatibility function of the query with the corresponding key, as shown in equation (1) and (2).

$$\text{Attention}(Q, K, V) = \text{softmax}(A) \cdot V, \tag{1}$$

$$A = \frac{QK^T}{\sqrt{d_k}} \tag{2}$$

$A \in \mathbb{R}^{t \times t}$ is the matrix of attention score. The self-attention module is designed to to learn long-range dependencies between vertices along the streamline by capturing interactions across different timesteps. In this method, a multi-head self-attention mechanism is employed to enhance the model's ability to comprehensively represent information. To ensure consistency between the training and inference processes, the self-attention score matrix must be partially masked. Specifically, the computed attention matrix $A$ is masked along the diagonal as follows:

$$a^{ij} = \begin{cases} a^{ij}, i \geq j \\ 0, i < j \end{cases}$$

Thus, at any given timestep during training, the streamline can only attend to known vertices and cannot access unknown ones. At the bottom of this block, a fully connected layer is used to transform the 192-dimensional embeddings, into 3-dimensional prediction of propagation direction $\hat{y}_t$.

*2.3 Loss Function*

As shown in Eq. 3, the loss function is defined as a weighted mean square error,

$$WL = \frac{1}{NT}\sum_{i=1,S_i \in S}^{N} \beta_j (\sum_{t=1,y_t \in S_i}^{T} \| \hat{y}_t - y_t \|_2) \qquad (3)$$

where $N$ represents the number of all training streamlines. $\beta_j = \frac{exp\,(N_j)}{\sum_{j=1}^{J} exp\,(N_j)}$, indicates the proportion of the bundle class $j$ to which the streamline $S_i$ belongs, relative to all $J$ fiber bundle classes. This parameter adjusts the relative importance of the loss from streamlines belonging to different bundles. Without this weighted loss, we encountered class imbalance during training, as some tracts are inherently smaller and thus have fewer training streamlines, making them "hard-to-track" for our model. We address this issue through treating each bundle as a distinct category and amplify the losses of streamlines associated with smaller tracts to ensure they are given more emphasis during training.

## 3. Experiments

*3.1 Implementation Details*

Our experiments were conducted on NVIDIA RTX2080 Ti GPU using the PyTorch framework. The number of Transformer decoder blocks was 6, with 6 attention heads, 192 dimensions and a block size of 96, training batch size was 2048.

To fit the block size, all training streamlines were set to a fixed length of $T = 96$ and a fixed step size of 1 mm by truncating longer sequences or padding shorter sequences with zero vectors. Tracking processes were also performed at the fixed step size of 1mm.

*3.2 Experiment1: Performance on ISMRM 2015 Tractography Challenge*

To access the performance of our model and compare it to other state-of-the-art methods: Learn-to-Track(Poulin et al., 2017), Entrack(Wegmayr and Buhmann, 2021), Track-to-Learn(Théberge et al., 2021), Neher et al.(Neher et al., 2017), we conducted tests using the ISMRM2015 Challenge dataset (Maier-Hein et al., 2017). The ISMRM2015 Challenge dataset is a synthetic brain phantom designed to evaluate tractography algorithms. The final tractogram was generated by performing tracking on one Human Connectome Project (HCP) subject, manually segmenting streamlines into 25 bundle masks, and applying bundle-specific deterministic tracking within these masks. Based on the predefined ground truth streamlines, a synthetic DWI at 2 mm isotropic resolution and a T1w image at 1mm isotropic resolution were generated using Fiberfox (Neher et al., 2014) tool. Noise and artifacts were added to mimic real-world MRI conditions, providing a controlled yet challenging dataset for benchmarking algorithms.

We used 5 HCP subjects for training, consistent with the number used in previous studies(Neher et al., 2017; Poulin et al., 2017; Théberge et al., 2021; Wegmayr and Buhmann, 2021). Human Connectome Project (HCP) dataset (Van Essen et al., 2013; Van Essen and Ugurbil, 2012; Van Essen et al., 2012) consists of high angular resolution, multi-shell diffusion data that acquired at 1.25 mm isotropic spatial resolution for 270 directions distributed over 3 shells. Training samples were generated using the MRtrix software package(Tournier et al., 2019). For each subject, we applied multi-shell multi-tissue constrained spherical deconvolution (MSMT-CSD)(Jeurissen et al., 2014) algorithm to estimate the fODF. Based on the estimated fODF, the probabilistic iFOD2 tracking algorithm (Tournier et al., 2010) was employed to generate the reference tractogram with 200 000 streamlines, constrained to a length range between 20 and 200 mm.

For testing, we directly used the preprocessed ISMRM2015 Challenge dataset provided by Renauld et al.(Theaud et al., 2019). The preprocessing steps included brain extraction, denoising, eddy current correction, N4 bias field correction, cropping, normalization, and resampling, all performed using Tractoflow pipeline(Di Tommaso et al., 2017; Garyfallidis et al., 2018; Kurtzer et al., 2017; Theaud et al., 2019). Additionally, for our proposed method, the preprocessed ISMRM2015 data was further resampled to 1.25 mm resolution to match the resolution of training data for optimal performance. Tracking was performed at 7 seeds per voxel, using the same step size and the same length limitation as in the training process.

*Evaluation Metrics*

Tractometer (Côté et al., 2013) is used to evaluate our results on the ISMRM2015 dataset, providing both connectivity-level and image-level assessments, with the following metrics.

1. Valid Bundles (VB): The number of valid bundles that were correctly reconstructed and that exist in the ground-truth data.

2. Invalid Bundles (IB): The number of bundles that can be extracted from the reconstructed tractogram, but do not match any ground-truth bundle.

3. Valid Connections (VC): True positives. The percentage of streamlines belonging to valid bundles. It is replaced by Valid Streamlines (VS) in new version.

4. Invalid Connection (IC): The percentage of streamlines belonging to invalid bundles.

5. No Connections (NC): The percentage of streamlines early stops, failing to connect two cortex regions. Noting that, IC and NC are both include as Invalid Streamlines (IS, false positives) in new version.

6. Overlap (OL): Proportion of voxels simultaneously covered by the reconstructed streamlines and the ground truth bundles.

7. Overreach (OR): Proportion of voxels covered by the reconstructed streamlines but not covered by ground truth bundles.

8. F1 (Dice) score: Overall similarity between the reconstructed streamlines and the ground truth, reflecting both precision and recall.

While Tractometer now offers updated metrics, we used the older version for comparisons with DL-based algorithms(Neher et al., 2017; Poulin et al., 2017; Théberge et al., 2021; Wegmayr and Buhmann, 2021) and the updated version for PFT(Girard et al., 2014) to ensure consistency with their reported results. Notably, Tract-to-learn (Théberge et al., 2021) can only be trained on a single subject, here we used the subject with ID 100206. Besides, the metrics for Learn-to-track (Poulin et al., 2017) were extracted from (Théberge et al., 2021), as the original paper did not provide similar experiment on the ISMRM2015 dataset.

*3.3 Experiment2: Performance on Tractoinferno dataset*

To explore the performance on in-vivo data, we trained and tested our model on Tractoinferno dataset. It's a high-quality and large-scale multi-site dataset (Poulin et al., 2022), consisting of 3T dMRI data at 1 mm isotropic resolution collected from 6 sites under varying acquisition protocols. This dataset includes 198 cases of training data, 58 cases of validation data and 28 cases of test data. Each subject is provided with single-shell data at b=1000s/mm² and a reference tractogram. For every subject, multiple tractography algorithms were employed to generate multiple tractograms. The final whole-brain reference tractogram was consolidated from them and was quality-controlled to represent "ground truth". Every tractogram contains 200 000 streamlines. In this work, 20 randomly selected subjects from the training set were used to train our model and all 28 subjects from test set were used to evaluate the performance of our model.

We trained the Learn-to-Track (Poulin et al., 2017) and Track-To-Learn (Théberge et al., 2021) for comparison. As in Experiment 1, Track-To-Learn was trained using only one subject (ID: 1001). The tracking parameters were kept consistent across Learn-to-Track (Poulin et al., 2017), Track-To-Learn (Théberge et al., 2021) and our method.

Tracking was seeded from the white matter with 5 seeds per voxel, using the same step size of 1 mm and identical length limitation of 200mm.

*Evaluation Metrics*

TractoEval (Di Tommaso et al., 2017; Garyfallidis et al., 2018; Kurtzer et al., 2017; Rheault, 2020) pipeline was used to evaluate the performance of our model on the Tractoinferno testing set. The pipeline provides three metrics: Dice, Overlap, Overreach.

*3.4 Experiment3: Ablation study*

In this experiment, we analyze the effectiveness of the proposed information in tractography. Baseline model consisted solely of MLP blocks. It received only local signals from streamline vertices as the input sequence. To validate the effectiveness of contextual information, a second model was implemented with attention mechanism but without the CNN layer. The neighbourhood information was then incorporated into the second model to create a third variant. Both of these models were trained using a mean squared loss function, without addressing the issue of bundle class imbalance. Finally, the fourth model introduced a weighted MSE loss function to mitigate this imbalance. All models were trained on same subjects and the tracking parameters were identical to Experiment 2 during tracking.

*3.5 Experiment4: Generalization ability on different sites*

To explore the generalization ability of our model, 20 subjects from site 1 were randomly selected from Tractoinferno training set and used to retrain the model. The model was then evaluated on all 28 subjects in the testing set, under two training scenarios: one trained exclusively on the 20 subjects from site 1 and the other trained on 20 subjects from all 6 sites, as in Experiment 2. The training and tracking parameters were same with Experiment 2.

## 4. Result

*4.1 Performance on ISMRM 2015 Tractography Challenge*

We present the metric results of our method on ISMRM2015 dataset compared to deep learning-based tractography methods in **Table 1**. Compared to Neher et al. (Neher et al., 2017), Learn-to-Track (Poulin et al., 2017) and Entrack (Wegmayr and Buhmann, 2021), our method present higher VC, VB and competitive Overlap rates. When compared with Track-to-Learn (Théberge et al., 2021), we report more valid bundles. The "hard-to-track" anterior commissures (CA) was reconstructed by our method, even though we present a lightly lower VC rate of it.

**Fig. 2** shows a visual comparison between the reconstructed and ground-truth tractograms. The proposed method well reconstructs the medium-difficulty bundle with a Dice above 0.6 and closely matches the shape of the hard-to-track bundle (Dice=0.567). For the very hard-to-track bundle, the reconstructed streamlines connect the correct gray matter regions, though the Dice coefficient is below 50%.

**Table 2** compares the performance of our method and PFT method using new version evaluation tool. When both seeding in white matter region the proposed method achieves higher dice scores and a higher VS in comparison with PFT(Girard et al., 2014).

**Table 1**
Results on ISMRM2015 Challenge dataset. Bold text indicates the best-performing result. Underlined text indicates sub-optimal results. Scores of Neher et al. (Neher et al., 2017), Learn-To-Track (Poulin et al., 2017) and Track-To-Learn(Théberge et al., 2021) are extracted from the paper by Track-To-Learn(Théberge et al., 2021). "Learn-to-Track (fODF)" indicates the method uses fODF data for network inputs.

|  | VC (%↑) | VB (↑) | IC (%↓) | NC (%↓) |
|---|---|---|---|---|
| Neher et al. | 52 | 23 | - | - |
| Learn-to-Track (fODF) | 49.8 | 23 | 43.1 | 7.1 |
| Learn-to-Track | 27.5 | **24** | 55.8 | 16.7 |
| Entrack | 52 | **24** | - | - |
| Track-To-Learn (SAC) | **68.5** | 23 | 30.7 | **0.8** |
| Proposed | 66.2 | **24** | **28.3** | 5.5 |
|  | IB (↓) | Dice (↑) | OL (%↑) | OR (%↓) |
| Neher et al. | **94** | - | 59 | **37** |
| Learn-to-Track (fODF) | 172 | - | 65.6 | - |
| Learn-to-Track | 218 | - | 56.2 | - |
| Entrack | 123 | 54 | 58 | 39 |
| Track-To-Learn (SAC) | 186 | - | **65.8** | - |
| Proposed | 136 | **56.6** | 63.8 | 42.3 |

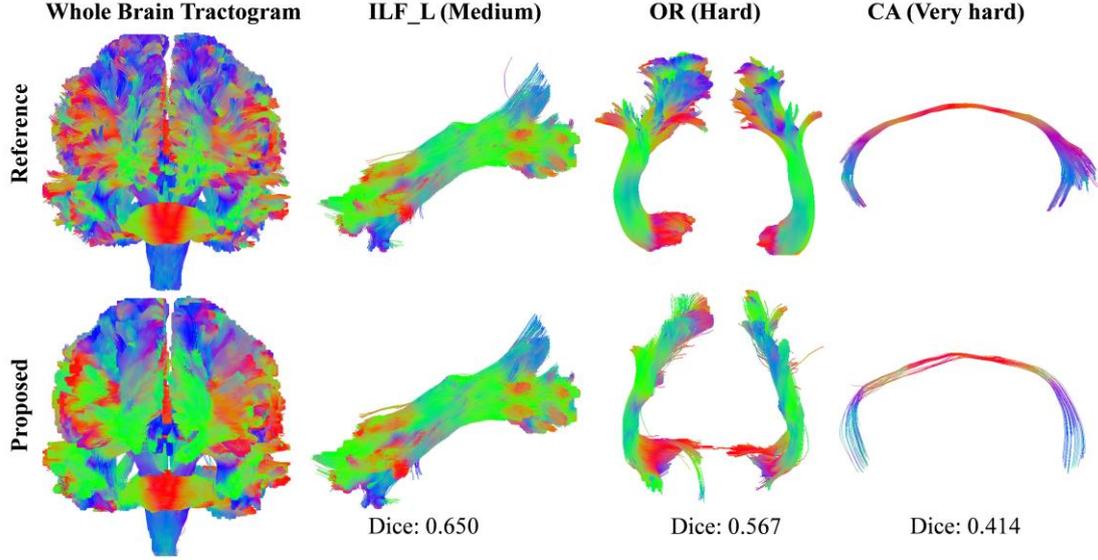

**Fig. 2.** Visualization of the tractogram from the ISMRM 2015 Challenge dataset, including the whole-brain tractogram and three bundles: left inferior longitudinal fasciculus (ILF_L), optic radiation (OR), anterior commissure (CA), categorized as *medium-*, *hard-*, and *very hard-to-track bundles*.

**Table 2**

Results on ISMRM2015 Challenge dataset, presented as mean (standard deviation). Bold text indicates the best-performing result. Underlined text indicates sub-optimal results. The results for local tracking and PFT tracking are extract from paper by (Renauld et al., 2023). "PFT tracking_1" indicates that the method sets seed points at the whole white matter region, "PFT tracking_2" the gray-white matter junction.

| Methods | VS (%↑) | VB (↑) | IS (%↓) | Total nb (↓) |
|---|---|---|---|---|
| local tracking | <u>42.0</u> | <u>19</u> | <u>58.0</u> | 4 332 527 |
| PFT tracking_1 | 33.5 | <u>19</u> | 66.5 | 2 203 575 |
| PFT tracking_2 | 28.8 | <u>19</u> | 71.2 | **1 002 726** |
| Proposed | **64.4** | **20** | **35.6** | <u>1 267 356</u> |
| | Dice (↑) | OL (%↑) | OR (%↑) | |
| local tracking | 51.7 | **82.6** | 121.3 | |
| PFT tracking_1 | 58.2 | <u>68.7</u> | 54.1 | |
| PFT tracking_2 | <u>58.4</u> | 64.9 | **44.5** | |
| Proposed | **59.2** | 65.2 | <u>48.3</u> | |

*4.2 Performance on Tractoinferno dataset*

**Table 3** presents the evaluation metric results on Tractoinferno testing set, comparing our method with state-of-the-art methods. Among the evaluated methods, PFT achieves the highest Dice and Overlap scores, with our method ranking second. However, PFT exhibits a higher Overreach compared to our approach. In addition, Track-to-Learn (Théberge et al., 2021) shows an exceptionally low overreach (5.4%) but suffers from the lowest Dice score (48%), likely due to its inability to generate a sufficient number of valid streamlines. In contrast, our proposed method effectively balances overreach and overlap, achieving competitive results.

**Fig. 3** shows the reconstruction of seven tracts that is of medium and hard difficulty to track, by referring to the paper by Maier-Hein et al (Maier-Hein et al., 2017). These tracts include the corpus callosum (frontal lobe most anterior part, CC_Fr_1), the left and right pyramidal tracts (PYT), the left and right arcuate fasciculus (AF_L and AF_R), the left and right Optic radiation and Meyer's loop. As observed, the hard-to-track bundles, PYT, OR_ML, CC_Fr_1 reconstructed by our method exhibit the higher similarity with ground truth bundles than Learn-To-Track (Poulin et al., 2017) and Track-to-Learn (Théberge et al., 2021). Also, our method induced the smallest number of false-positive streamlines of the bundle (AF_R). **Fig. 4** presents a scatter plot of Dice scores across 28 testing cases for Learn-to-Track, Track-to-Learn, and the proposed method. As seen in **Fig. 3**, the proposed method demonstrates more consistent results across subjects and tracts. Especially for the hard-to-track AF and OR_ML tracts, the number of failed cases (i.e., Dice = 0) is significantly lower. This highlights the robustness of the proposed method in handling challenging tracking scenarios.

In **Fig. 5**, we visualize the attention scores across all of the model's layers and heads computed during the reconstruction of one particular streamline. Attention maps are presented in tabular form, with rows representing layers and columns representing heads. It conveys the coarse shape of the attention pattern and enables us to see how attention patterns evolve throughout the model. The vertical axis of each attention graph indicates the current position of the tracking, and the horizontal axis indicates the positions to which the model is attending. As can be seen from the graphs, the network does not focus on the unknown location when predicting at each time step t. Initial layers tend to be position-based, i.e. focusing on the current position (layer0, heads 0,1,3-5) or focusing on both the current position and its adjacent previous position (layer0, head2). In contrast, deeper layers tend to capture a wider range of information, with different heads exhibiting distinct attention patterns.

**Table 3**

Tractography evaluation measures across 28 subjects, presented as mean (standard deviation). Bold text indicates the optimal scores.

|  | Dice | Overlap | Overreach |
| --- | --- | --- | --- |
| Learn-To-Track (DWI) | 58.0(3.2) | 53.3(3.9) | 25.8(6.7) |
| Track-To-Learn (SAC) | 48.4(2.9) | 35.1(2.3) | **5.4(1.5)** |
| Proposed | <u>64.5(2.2)</u> | <u>59.0(2.9)</u> | <u>21.7(6.1)</u> |
| PFT | **69.1(7.8)** | **69.2(12)** | 25.7(6.2) |

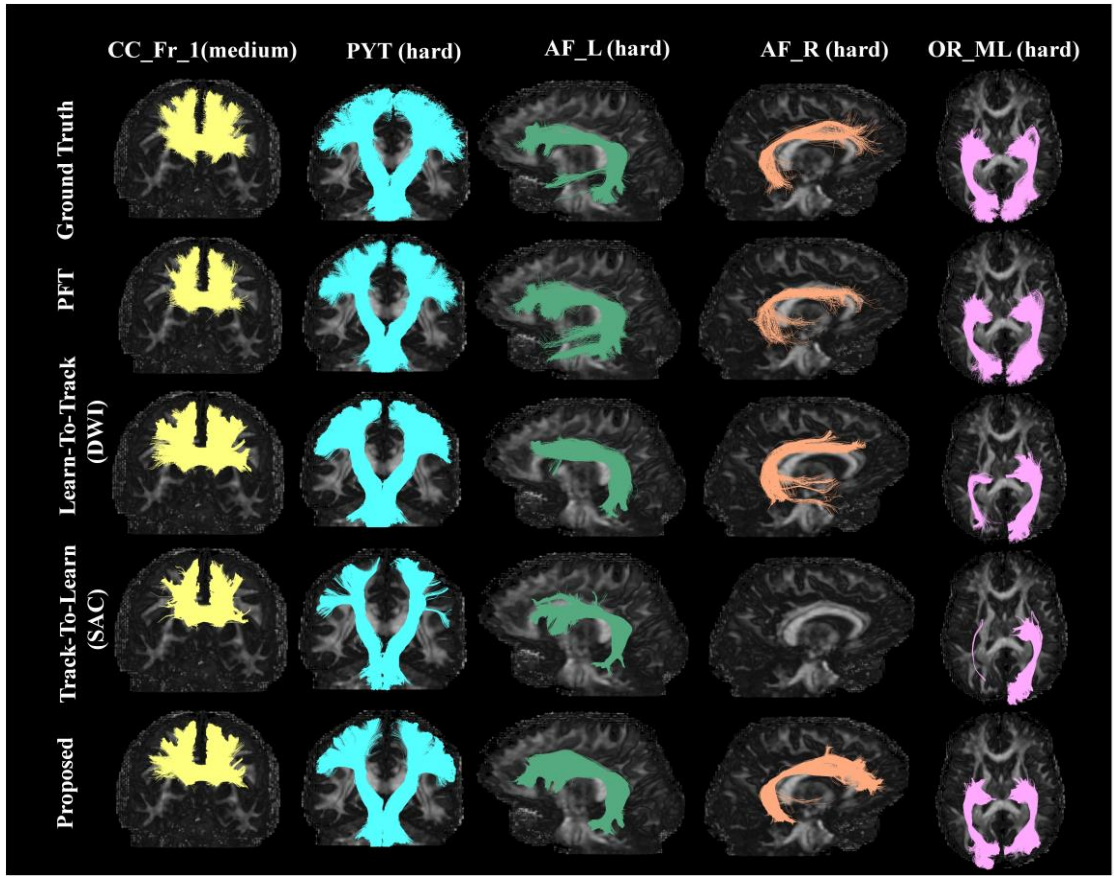

**Fig. 3.** Visualization of the reconstructed bundles from sub-1006. The left and right pyramidal tract (PYT_L and PYT_R) are displayed together in one segment, same as the Optic radiation and Meyer's loop (OR_ML_L and OR_ML_R)

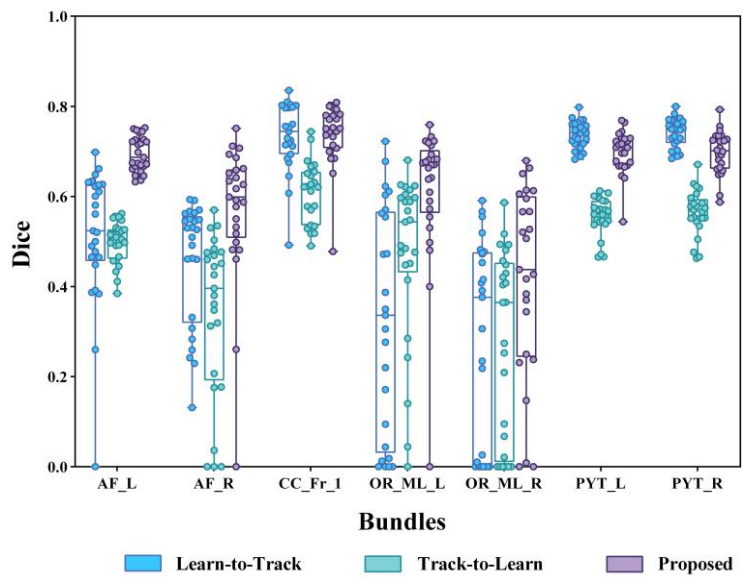

**Fig. 4.** Dice scores for the seven tracts across all Tractoinferno testing subjects, obtained using

different methods: *Learn-to-Track*, *Track-to-Learn*, and the *Proposed* method.

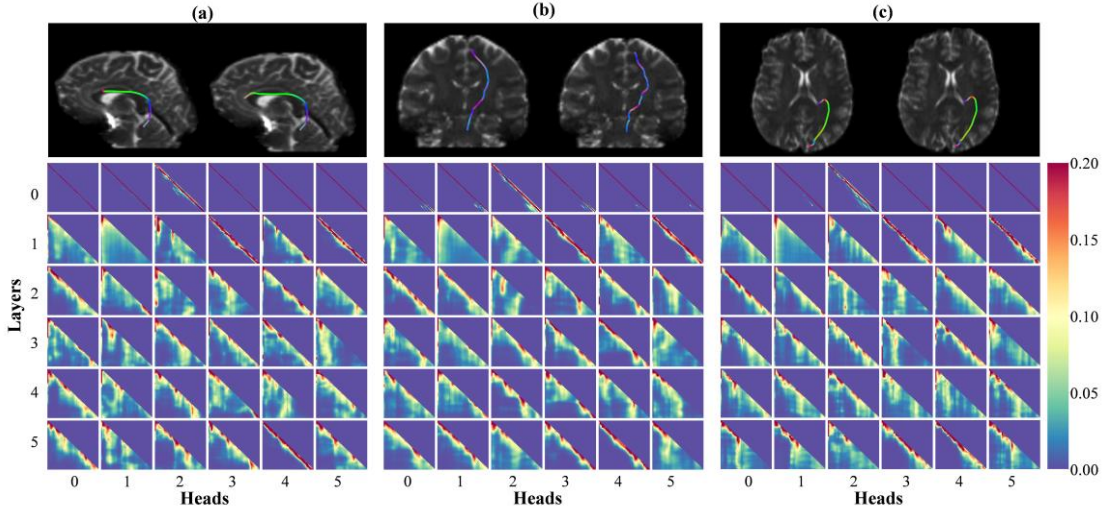

**Fig. 5.** Attention scores of three reconstructed streamlines (left) and corresponding references (right), that belong to (a) AF_L, (b) PYT_L, (c) OR_ML_L tracts.

*4.3 Ablation study*

We report the impact of different information on performance in **Table 4** and visualize one reconstruction result (sub-1006) in **Fig. 6**. The results demonstrate that a model lacking novel information fails to successfully reconstruct the whole-brain tractogram. The incorporation of contextual information enhances the integrity of the reconstructed streamlines. Furthermore, with both neighbourhood and contextual information aggregated in the procedure, the model becomes able to predict more accurate propagation directions. Finally, as the importance of different bundles being scaled during training, the Overlap rates of the whole-brain tractogram, i.e. white matter coverage, increases and the Overreach rate slightly suffers, which can be seen within the black short dotted box. As a result, the dice score shows improvement compared to models that do not employ the balanced loss function.

**Table 4**
Ablation studies of our proposed method. Results are presented as mean (standard deviation).

| Contextual Information | Spatial Information | Class Imbalance Information | Dice | Overlap | Overreach |
|---|---|---|---|---|---|
| × | × | × | 0.8(0.5) | 0.4(0.3) | **1.9(1.8)** |
| √ | × | × | 22.8(3.9) | 15.8(3.1) | 9.6(2.9) |
| √ | √ | × | 62.7(3.4) | 56.1(3.7) | 19.2(3.6) |
| √ | √ | √ | **64.5(2.2)** | **59.0(2.9)** | 21.7(6.1) |

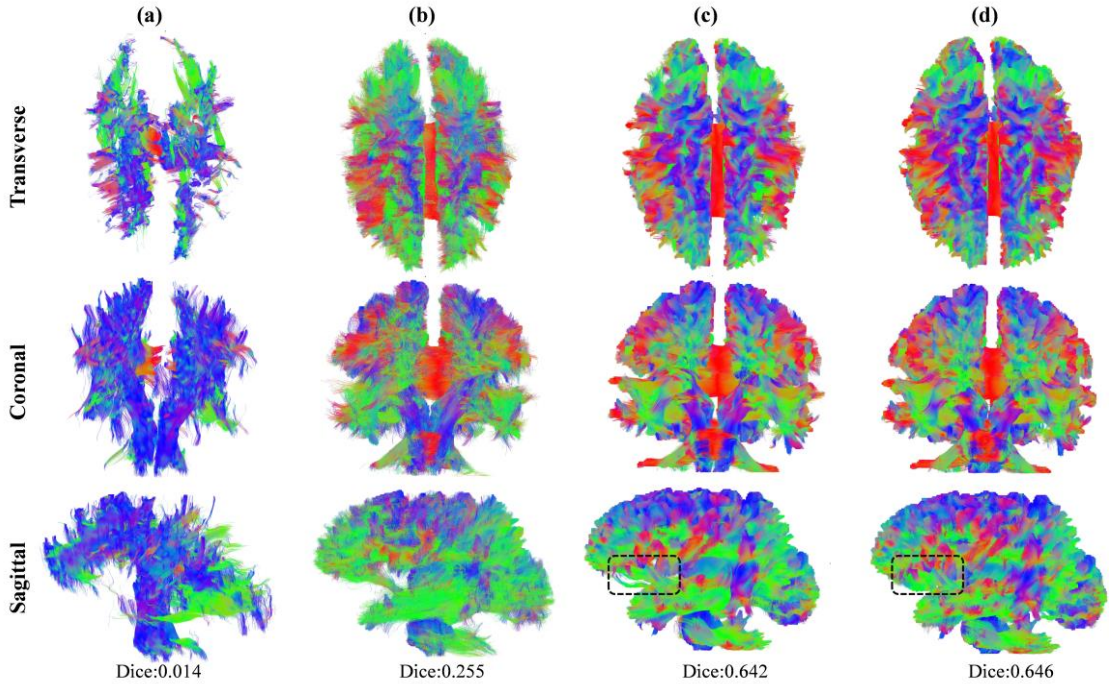

**Fig. 6.** Reconstructions of sub-1006 with different information incorporated during tracking. Each column, from left to right, shows the results of four models. These models progressively integrate: (a) no novel information, (b) contextual information, (c) spatial information, (d) class imbalance information. The corresponding Dice scores are displayed at the bottom of each column.

*4.4 Generalization Capabilities on different sites*

**Table 5** compares the test results of our model trained with the same amount of data, sourced either from all six sites (Sites 1-6) or exclusively from Site 1. The table includes the overall test results across all test data, the results specifically for subjects from Site 1 in the test set, and the results for subjects from the remaining sites (Sites 2–6). As observed, the single-center training data does not significantly reduce the test performance compared to the multi-center training data, indicating the generalization abilities on unseen dataset.

**Table 5**

Comparison of testing results from models trained by multi-center and single-center training data. Results are presented as mean (standard deviation).

|  | Training | | | | | |
|---|---|---|---|---|---|---|
|  | All sites (20 subs) | | | Site1 (20 subs) | | |
|  | Dice | Overlap | Overreach | Dice | Overlap | Overreach |
| All sites (28 subs) | 64.5(2.2) | 59.0(2.9) | 21.7(6.1) | 62.7(2.8) | 56.1(3.2) | 19.3(3.3) |
| Site1 (8 subs) | 63.7(1.5) | **60.0(2.8)** | 25.8(7.6) | 62.0(2.7) | 56.7(3.0) | 21.2(3.2) |
| Sites2-6 (20 subs) | **64.9(2.5)** | 58.4(2.9) | 19.4(3.6) | 63.1(2.9) | 55.8(3.4) | **18.2(3.0)** |

## 5. Discussion

Diffusion MRI-based tractography enables the non-invasive reconstruction of white matter pathways in the brain. However, the poor performance of tractography, characterized by high rates of false positives and false negatives, remain a persistent challenge. One way to reduce false-positives is to launch millions of seeds to reconstruct a tractogram and then employ advanced filtering techniques, such as SIFT, LIFE, COMMIT (Daducci et al., 2015; Pestilli et al., 2014; Schiavi et al., 2020; Smith et al., 2015) to filter false positives. But this comes with computational cost, as well as impacts on construction results, such as laterality changes in brain structural connectivity(He et al., 2024). And after filtering, false-negatives challenge remains, i.e., the spatial coverage of reconstructed beams passing through "hard-to-track" regions remains low. This issue is particularly pronounced in regions with complex fiber configurations, where accurate reconstruction is difficult when relying solely on limited local information. In this study, we present a deep learning-based tractography algorithm incorporating novel information to address this challenge and evaluate its reliability compared to state-of-the-art DL-based tractography methods—namely Learn-To-Track(Poulin et al., 2017) , Entrack(Wegmayr and Buhmann, 2021), DeepTract(Benou and Riklin Raviv, 2019), Track-to-Learn(Théberge et al., 2021) and the approach by Neher et al. (Neher et al., 2017). Unlike these DL methods, which explicitly append directional information from a limited number of proximal streamline segments as input, our method offers a more flexible framework. It leverages spatial neighborhood and contextual information stored in diffusion MRI signals from known streamline segments available for prediction, leading to improved fiber reconstruction, even in hard-to-track regions where advanced DL-based methods, such as Learn-To-Track(Poulin et al., 2017) and Track-to-Learn(Théberge et al., 2021), face difficulties.

The promising performance of our method stems from the effective combination of both spatial neighbourhood information and fiber pathway contextual information through the convolutional layer and Transformer-decoder blocks. By incorporating cues from surrounding voxels through a CNN model, our method enhances local structural coherence, improving its ability to resolve multiple fiber configurations within a voxel. However, resolving local ambiguities alone is insufficient for accurate fiber tracking. Predicting the correct propagation direction of fibers requires consideration of streamline-specific context. To achieve this, we employ a Transformer-decoder block that leverages the attention mechanism to capture long-range relationships across the entire streamline. This enables the model to identify the correct direction within a voxel by aligning predictions with the global fiber pathway. Together, the integration of spatial information and streamline context provides a complementary framework that improves local accuracy of fibers, while ensuring global coherence along entire fiber pathways. This is evidenced in Experiments 1 and 2, where our method exhibited competitive performance with a relatively high VC rate, a high number of reconstructed VB, and a high Overlap rate, as well as significant improvements in resolving complex fiber patterns.

Instead of directly inputting raw diffusion MRI signals into the network, we first transform them into a fixed-order SH representation. This strategy allows our method to handle data acquired with different schemes, including variations in the number of gradient directions and b-values. Additionally, the SH representation improves robustness to noise and facilitates data harmonization across centers(Chen et al., 2025). A 6th-order SH representation is sufficient to accurately capture diffusion signals while keeping the parameter count manageable. For a reliable SH representation, at least 28 DW volumes with uniformly distributed directions are recommended, which is feasible in clinical settings.

However, the proposed method has its limitations. While it effectively leverages spatial neighborhood and contextual information, other potentially valuable cues, such as the individual locations of gray matter and white matter boundaries, remain unutilized. In its current implementation, there are still hard-to-track bundles that cannot be correctly reconstructed, which urges us to utilize more information in further(Calixto et al., 2024). Moreover, the fixed step size used during tracking, combined with first-order integration, may cause overshooting in high-curvature regions, redundant iterations and accumulation of tracking errors. Same with most iterative tracking methods, the high-quality reconstruction is time-consuming. Future efforts will focus on accelerating tractography through lightweighting the model, optimizing step size adaptively, or employing second-order integration methods to improve both efficiency and accuracy.

Overall, we have demonstrated that neighbourhood and contextual information from diffusion signal can be used to enhance tractography. Other types of novel information, such as the logic that the real tissue use to determine its pathways, may provide cues for us to rethink tractography(Schilling et al., 2019a).

# 6. Conclusion

In this work, we proposed a deep-learning-based tractography algorithm that integrates multiple sources of information, including image-domain neighborhood features, global context along the streamline, and class imbalance of training samples. Our method successfully generated anatomically plausible streamlines across both synthetic and in-vivo human brain datasets, delivering high-performance reconstructions, particularly in hard-to-track regions. These findings underscore the importance of integrating novel information to enhance tractography and pave the way for future efforts to further improve the accuracy, robustness, and anatomical reliability of white matter mapping.

## CRediT authorship contribution statement

**Yiqiong Yang:** Writing – original draft, Visualization, Validation, Methodology. **Yitian Yuan:** Writing – review & editing. **Baoxing Ren:** Writing – review & editing. **Ye Wu:** Writing – review & editing. **Xinyuan Zhang:** Writing – review & editing, Methodology, Validation, Supervision, Conceptualization.

## Declaration of competing interest

The authors declare that they have no known competing financial interests or personal relationships that could have appeared to influence the work reported in this paper.

## Acknowledgments

This work was supported by the National Natural Science Foundation of China ( U21A6005, 61971214, 82372079), the Guangdong Basic and Applied Basic Research